\documentclass[letterpaper, 10 pt, conference]{ieeeconf}
\IEEEoverridecommandlockouts 
\usepackage{graphicx}
\usepackage{float}
\usepackage{hyperref}
\usepackage[ruled,vlined]{algorithm2e}
\usepackage{algorithmic}
\usepackage{amsmath,amssymb,amsfonts}
\usepackage{svg}
\usepackage{multirow}

\newcommand{\etal}{\textit{et al.}}
\hypersetup{
    colorlinks=true,
    linkcolor=blue,
    urlcolor=cyan,
    }
\title{\LARGE \bf
Sim2Real Instance-Level Style Transfer for 6D Pose Estimation
}
\author{Takuya Ikeda, Suomi Tanishige, Ayako Amma, Michael Sudano, Hervé Audren, and Koichi Nishiwaki$^{*}$% <-this % stops a space
\thanks{This work has been submitted to the IEEE for possible publication. Copyright may be transferred without notice, after which this version may no longer be accessible.}%
\thanks{$^{*}$ All authors are with the Woven Planet Holdings, Inc. 3 Chome-2-1 Nihonbashimuromachi, Chuo City, Tokyo 103-0022, Japan, {\tt\small [firstname.lastname]@woven-planet.global}}%
}

\begin{document}

\maketitle
\thispagestyle{empty}
\pagestyle{empty}

\begin{abstract}

In recent years, synthetic data has been widely used in the training of 6D pose estimation networks, in part because it automatically provides perfect annotation at low cost. However, there are still non-trivial domain gaps, such as differences in textures/materials, between synthetic and real data. These gaps have a measurable impact on performance. 
To solve this problem, we introduce a simulation to reality (sim2real) instance-level style transfer for 6D pose estimation network training.
% an unsupervised domain-adaptation method for 6D pose estimation via style transfer. 
Our approach transfers the style of target objects individually, from synthetic to real, without human intervention. 
% Our approach transfers the style of real target objects into simulation individually without human intervention. 
This improves the quality of synthetic data for training pose estimation networks. We also propose a complete pipeline from data collection to the training of a pose estimation network and conduct extensive evaluation on a real-world robotic platform. 
% Our evaluation shows the improvement in both pose estimation performance and the quality of images adapted by the style transfer.
Our evaluation shows significant improvement achieved by our method in both pose estimation performance and the realism of images adapted by the style transfer.

\end{abstract}

% TODO(taku): unification of terminology. e.g. sim, synthetic
% TODO(taku): we -> passive voice
% TODO(taku): texture-rich, content-rich
\section{Introduction}

% Recently, the use of service robots in smart spaces, such as houses and cities, has been gaining interest. Such robots are expected to perform tasks that require object recognition and manipulation. In order to carry out these services, the estimation of both position and orientation (6D pose) of objects in a scene is essential.

% 6D pose estimation, 
Determining the location and orientation (6D pose) of detected objects in the 3D world, is a critical ability for robots that are meant to manipulate objects. As the field progresses towards robot deployment in increasingly unstructured settings, such as human homes, the ability to train robust perception systems 
% capable of 6D pose estimation 
across a wide variety of situations has become a key bottleneck.
Complicating matters, 
not only is there a large variety of household objects, %consumer packaged goods (CPGs) in everyday life,
but such objects can be transparent, shiny, or have any number of challenging characteristics. % complicating characteristics.
% Among the numerous available vision sensors, RGB cameras have the desired characteristics of being low cost and having the ability to acquire visual information as humans do.
In this paper, we tackle RGB-based 6D pose estimation for these types of challenging objects.
% We target to use a RGB image for estimating the 6D pose of objects.

% Deep neural networks have been greatly improved in RGB-based 6D pose estimation \cite{Rad2017-jb, Kehl2017-bf, xiang2017posecnn}.
% Deep neural network-based 6D pose estimation has made great progress in recent years \cite{Rad2017-jb, Kehl2017-bf, xiang2017posecnn, hashimoto2020kosnet}.
Deep neural networks have made great progress in RGB-based 6D pose estimation \cite{Rad2017-jb, Kehl2017-bf, xiang2017posecnn}.
% However, a large amount of data collection and data annotation are required in the training phase to achieve high performance; 
However, a large amount of data collection and annotation is required in the training phase to achieve high performance. Moreover, since annotation of 6D object poses is more difficult than 2D image labeling, the labor required for these tasks is immense, especially when only RGB images are available \cite{Hinterstoisser2019-cb, yang2021dsc}. 
% However, since the network is required to simultaneously classify and detect the objects, and estimate their poses, the pose estimation task is more data hungry compared to some of the other vision tasks such as image classification, object detection, etc \cite{yang2021dsc}; the labor required for the data annotation is immense \cite{Hinterstoisser2019-cb}. 
One way to overcome this difficulty is to leverage synthetic data; %, especially when 3D models are readily available; 
virtually infinite data and perfect annotations can be generated at low cost via rendering. However, although these features are attractive, there exists a distinct domain gap between synthetic and real data, resulting in sub-optimal network performance. 

\begin{figure}[htbp]
\centering
\includegraphics[scale=0.39]{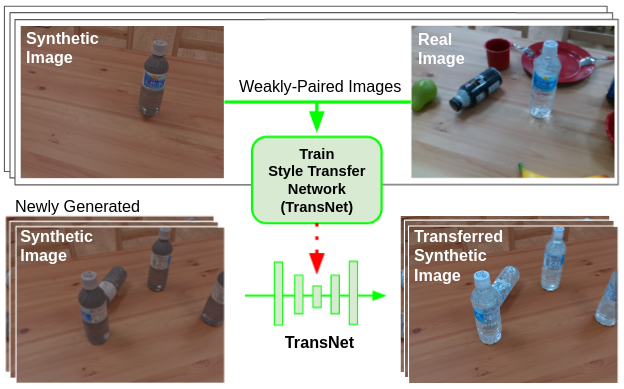}
\caption{Top: Training of the style transfer network using weakly-paired images between synthetic and real.  Bottom: Style transfer from newly generated synthetic images to real ones, using the style transfer network. %(a) Training of style transfer network using weakly-paired images between synthetic and real. (b) Style transfer from newly generated synthetic images to real ones, using the style transfer network.
}
\label{fig:motivation}
\end{figure}
In the past few years, numerous methods have been proposed to solve the domain gap problem. One promising approach is the combination of domain randomized (DR) and photo-realistic data, as described in \cite{tremblay2018deep, alghonaim2021benchmarking}.
% While these approaches generally have great results, they cannot guarantee such performance if the quality of 3D models or renderers are poor.
While these approaches have shown great promise when high-quality 3D models are available, % to facilitate high-fidelity rendering,
performance often degrades when using noisy 3D assets that are acquired in the wild.
Even when high-quality models and renderers are available, there still exists a gap between simulation and real.
To bridge this gap, domain adaptation techniques are promising, as they allow the conversion of synthetic images into realistic ones, utilizing real images without annotation. One popular approach here is the use of Generative Adversarial Networks (GANs). For example, Rojtberg \etal  ~\cite{Rojtberg2020-af} leveraged CycleGAN \cite{Zhu2017-mq} to transfer synthetic images to realistic ones for 6D pose estimation networks. However, as described in INIT \cite{Shen2019-jy}, if the target domain is a complex scene containing multiple objects, serious inconsistencies will occur because these methods \cite{Zhu2017-mq, huang2018multimodal, lee2020drit++} focus on directly adapting a global style to the entire image.
% Note(taku): 現実世界のものをすべてシミュレーションに再現できない場合も不一致が生じる

Our key insight is that transferring the style of individual objects instead of the whole image style will result in more correspondences between the domains.
% Moreover, pose-level correspondences of target objects are an important part of style transfer for multicolored objects;
Moreover, pose-level correspondences of multicolored objects are an important part of style transfer;
% non-uniformly colored objects; 
correspondences of instance locations in 2D space only will result in appearance mismatches of texture-rich objects.
For example %CPGs 
household objects often have a very different style depending on the viewpoint.
% Thus, in this paper, we propose a sim2real instance-level style transfer for 6D pose estimation networks. % an unsupervised domain adaptation method for 6D pose estimation via instance-level style transfer. 
Thus, we propose a sim2real domain adaptation method for 6D pose estimation via instance-level style transfer that is capable of handling complex scenes without manual annotations.
Our objective is to improve the capability of an off-the-shelf 6D pose estimation network by improving synthetic data quality. % without manual labor.
The proposed method is able to transfer multiple texture/material-rich object styles individually. 
% Our style transfer method extends single image translation \cite{Park2020-lf} to being able to handle instance-level correspondences on multiple images automatically.
% Our objective is to improve the capability of an off-the-shelf 6D pose estimation network by improving synthetic data quality without manual labor.
% To achieve pose-level correspondences, we generate weakly-paired images between simulation and real using a robot and a 6D pose estimation network trained on the synthetic data only.
% because texture and pose, especially orientation should be entangled for texture-rich objects 
\autoref{fig:motivation} shows an overview of our sim2real instance-level style transfer. 

To summarize, our key contributions are:
% \begin{enumerate}
%     \item \label{contrib:datagen} An entirely automatic pipeline to produce weakly-paired synthetic data for training sim2real domain-transfer network as described in \autoref{method}. 
%     \item \label{contrib:da} A general procedure for adapting synthetic images to target domain images that improves a 6D pose estimation network as demonstrated in \autoref{experiments}.
% \end{enumerate} 
\begin{enumerate}
    \item \label{contrib:datagen} An automatic pipeline to produce weakly-paired images between synthetic and real for training sim2real style transfer networks as described in \autoref{weakly-paired}. % as described in \autoref{method}.
    \item \label{contrib:da} A procedure for instance-level object style transfer from synthetic to realistic as described in \autoref{style_transfer}. % for adapting synthetic images to target domain images. % that improves the performance of a 6D pose estimation network as demonstrated in \autoref{experiments}.
\end{enumerate} 
Our contributions improve the performance of a 6D pose estimation network as demonstrated in \autoref{experiments}.

\section{Related Work}
\subsection{Learning with Synthetic Data}
% Kuni's comment: if there would be an extra space available in the paper, it's worth mentioning DatasetGAN as an interesting way to utilize synthetic data: https://nv-tlabs.github.io/datasetGAN/
Recently, synthetic data has been widely used in the training of 6D pose estimation systems because it naturally gives perfect annotations at a low cost.
% As mentioned in DOPE \cite{tremblay2018deep}, the combination of DR and photo-realistic rendered images is a key for generating highly-capable synthetic data for successful training of 6D pose estimation networks. 
DOPE \cite{tremblay2018deep} have successfully trained 6D pose estimation network using only synthetic data which contain DR and photo-realistic rendered images.
% DOPE \cite{tremblay2018deep} succeeded the capable 6D pose estimation network training by using the combination of DR and photo-realistic rendered images.
% Subsequently, they show great results for 6D pose estimation using only synthetic data.
For DR, several methods \cite{tremblay2018deep, tobin2017domain, Tremblay2018-ld, Sundermeyer2018-hx}\ are proposed. They randomize several parameters to get domain invariant features, such as poses of target objects and cameras, types of distractors, textures/materials of foreground and background, lighting settings, and sensor noise. While these methods show great results, % with the proper settings,
it requires domain knowledge to find good parameters. % to achieve high performance.
% there is some performance variety based on their knowledge of the target domain.
% Moreover, to cover the target domain distribution, the data tend to be the oversized and redundant.
With regards to creating photo-realistic images, there are also lots of research efforts \cite{alghonaim2021benchmarking, denninger2019blenderproc, roberts:2021, schwarz2020stillleben}. These sources prove that the qualities of the object model, such as textures/materials, and the renderers, including shadowing/lighting, are all important for sim2real. % the renderers, including shadowing, lighting, and anti-aliasing, are all important for sim2real.
However, even with high quality 3D models, there always exist a difference between rendered and real images; this is the domain gap that degrades performance of pose estimation.

\begin{figure*}[!ht]
\centering
\includegraphics[scale=0.28]{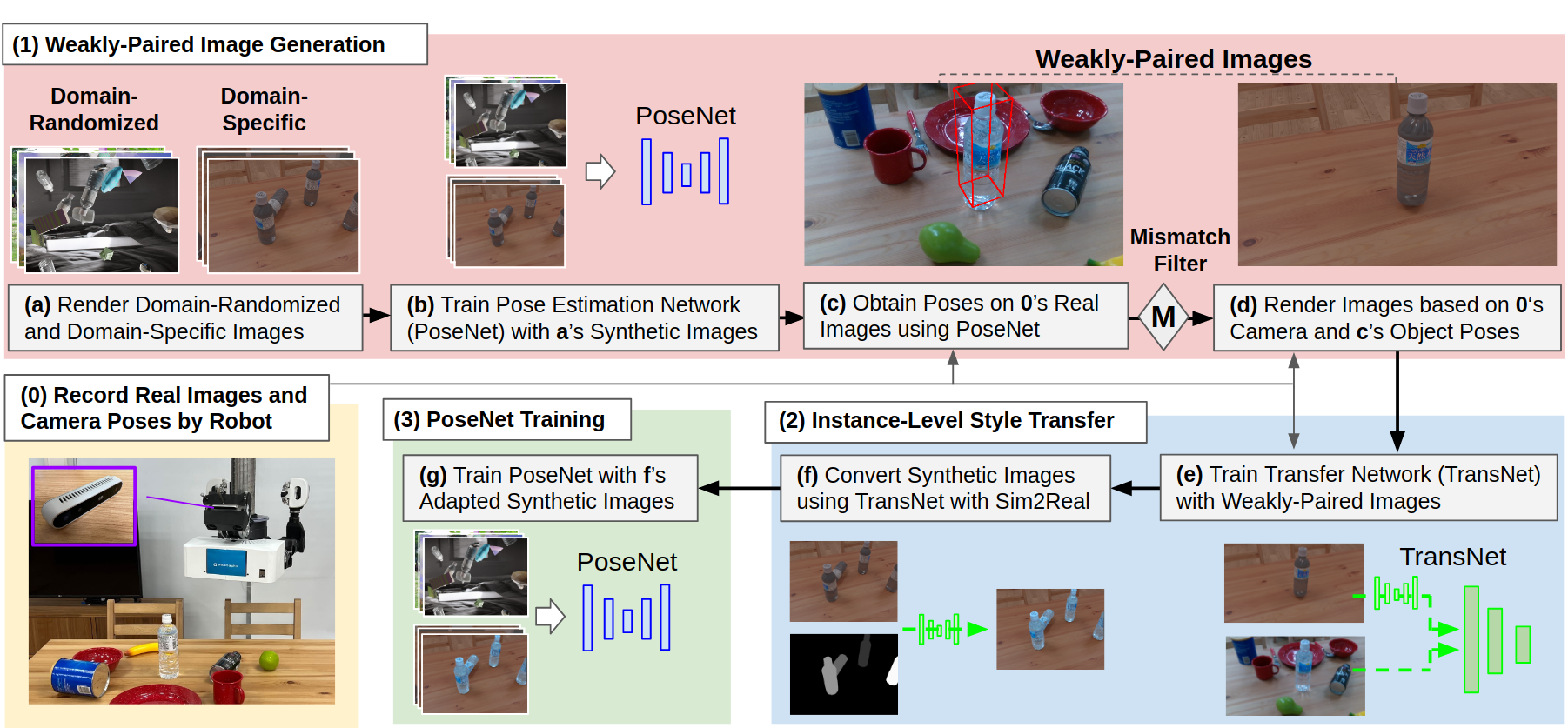}
\caption{\textbf{The workflow of the proposed domain adaptation and training method:} % $\rm(\,I\,)$ 
(1) Weakly-paired images are generated by using a robot and a 6D pose estimation network trained on synthetic images only. To generate these images properly, a mismatch filter is utilized as described in \autoref{weakly-paired}. % false-positive filter is utilized.
% $\rm(I\hspace{-.01em}I)$ 
(2) Then, a style transfer network is trained based on the weakly-paired images, and convert synthetic images via instance-level style transfer. % $\rm(I\hspace{-.15em}I\hspace{-.15em}I)$ 
(3) Lastly, the 6D pose estimation network is trained using the adapted synthetic images.}
\label{fig:sim2real_overview}
\end{figure*}

\subsection{Domain Adaptation}
% Instead of trying to precisely model the texture/materials of target objects, unsupervised domain adaptation methods seek to improve the quality of rendered images using real images.
Instead of trying to precisely model the textures/materials of target objects, domain adaptation methods seek to bridge these gaps using target domain's data \cite{bi2019deep, murez2018image}.
% When the target domain's data is available, unsupervised domain adaptation directly adapts the network to the target domain \cite{bi2019deep}. 
% This approach can greatly reduce manual operations and data redundancy, as mentioned previously. 
This approach can greatly reduce manual operations as mentioned previously. 
There are two primary methods of domain adaptation for 6D pose estimation networks without manual annotations. 
One way is to adapt the network to the target domain using geometric information. The reason for doing this is that geometry information is generally more domain invariant than RGB features and, therefore, a useful feature for pose estimation. For example, Self6D \cite{wang2020self6d} compares rendered depth and real depth in order to minimize the error using a differentiable renderer. 
Rad \etal ~\cite{Rad2019-af} also use depth information to enforce network invariant between synthetic and real. However, these methods are not available when depth information is lacking, as with transparent objects.
The other method of domain adaptation for 6D pose estimation networks is a GAN-based approach that can be applied using only RGB information. CycleGAN \cite{Zhu2017-mq} can translate unpaired images bidirectionally between two domains. For example, Rojtberg \etal ~\cite{Rojtberg2020-af} leverages the CycleGAN \cite{Zhu2017-mq} to get a mapping between synthetic and real domain for pose estimation. Moreover, DRIT \cite{lee2020drit++} and MUNIT \cite{huang2018multimodal} can handle multiple latent spaces for image-to-image translation. For example, they are able to disentangle style and content.
However, if the target scene is complicated, for example containing multiple colorful objects, % multiple discrete objects, or even if a single object is texture-rich or contains multiple colors, 
serious inconsistencies can occur. This is because these methods \cite{Zhu2017-mq, huang2018multimodal, lee2020drit++} focus on directly adapting a global style to the entire image.
Richter \etal ~\cite{richter2021enhancing} translates synthetic images into realistic ones by using lots of intermediate rendering features for training. % They show moderate improvements in performance but don't separate instance-level content from context, which is what we focus on. Moreover, since we only need one rendering feature which is instance-mask information, our approach is simpler since we only require instance masks for synthetic labels.
They show moderate improvements in performance by operating at the class level. By contrast, our approach is more targeted since we focus on individual instances and is simpler since we only require instance masks for synthetic labels.
INIT \cite{Shen2019-jy} achieves instance-level translation, but it requires 2D bounding box annotations of real data, while we don't require any manual annotations.

% We propose a sim2real domain adaptation method for 6D pose estimation via instance-level style transfer that is capable of handling complex scenes without manual annotations. The proposed method is able to transfer multiple texture-rich object styles individually. Our style transfer method extends single image translation \cite{Park2020-lf} to being able to handle instance-level correspondences on multiple images automatically.

\section{Method}  \label{method}
%Note(taku): リアルデータからはスタイルのみを抽出し、それをシミュレーションデータに使えるという趣旨をしっかり伝える
Our objective is to amplify the performance of 6D pose estimation networks (PoseNets) via style transfer, using synthetic and real data without manual annotations.
We hypothesize that less data is required for the training of style transfer networks (TransNets) than those for PoseNets. 
% conjecture
We first collect and use minimal real data for TransNet training. Then, transfer a large number of synthetic images to more realistic ones, for PoseNet training.
% In our case, each TransNet is trained with less than 1k images, and is then used to generate more than 2.5k images for PoseNet training.
Our approach improves TransNet performance by providing data in which pose and instance information coincides between synthetic and real domains. Since ground truth labels are only available on synthetic data, TransNet must maintain the object poses in the image during the transfer.
Our approach explicitly adapts the appearance of individual instances while keeping their poses in the scene.

To achieve the goal, our entire workflow is structured as shown in \autoref{fig:sim2real_overview}. 
First, we collect real images with a robot and generate synthetic images corresponding to these real images by using a PoseNet trained on the synthetic images only.
Then, TransNet is trained based on these image pairs in an unsupervised manner.
Finally, we train the PoseNet using the synthetic images adapted with the style transfer. 
In this section, we describe the key steps in detail.
We assume that 3D models of the environment and target objects are given, and the camera pose in the global frame is provided by the robot system.
% Note(taku): add the related comment in conclusion; this is reasonable for us, but... 

\subsection{Real Data Collection by Robot}  \label{data_collection}
Real images are required for our domain adaptation method, although annotations are not. We use the robot for data collection in the target environment, which is originally designed for carrying out services in the environment. At the same time, the poses of the camera in the environment coordinate system are stored. They are used for rendering % the corresponding images in later steps.
images which correspond to the collected real images in later steps.
During the data collection, the robot captures the same target area from multiple viewpoints. This constraint is introduced in order to decrease detection mismatches in later steps. % false-positives in later steps.
In our case, we utilized a gantry type robot installed on the ceiling of a room as shown in \autoref{fig:sim2real_overview} (0).

% $\rm(\hspace{.18em}i\hspace{.18em})$
% $\rm(\hspace{.08em}ii\hspace{.08em})$

\subsection{Weakly-Paired Images Generation}   \label{weakly-paired}
% For the latter instance-level TransNet training, 
% Our goal here is to render images of the target objects and environment which have a similar look to gathered real images, as shown in \autoref{fig:sim2real_overview} (c)(d). 
Our goal here is to render scenes including the target objects and environment, which look similar to our collected real images, as shown in \autoref{fig:sim2real_overview} (c)(d).
We call these \textit{weakly-paired images}.
% Our approach don't require the perfect pair between sim and real. 
Since we already have camera poses and 3D models, the only missing information needed for rendering is the 6D pose of the target objects. To get these poses, we train the PoseNet using only synthetic data. %  without adaptation.
Our approach can accept off-the-shelf RGB-based pose estimation networks.
To obtain good performance, we generate two types of synthetic dataset like DOPE \cite{tremblay2018deep}. 
One is the Domain-Randomized dataset generated under the condition of random-placement sampling without considering gravitational effect together with random background, distractors, and lights (see \autoref{fig:sim2real_overview} (a)); %'s \rm{\hspace{.18em}i\hspace{.18em}}); % $\alpha$); 
this aims to cover all the poses uniformly and realize robustness in variable conditions. 
The other one is the Domain-Specific dataset generated under the conditions of placing target objects considering gravitational effect in the 3D model of the target environment (see \autoref{fig:sim2real_overview} (a)); %'s \rm{\hspace{.08em}ii\hspace{.08em}}); %$\beta$);
this aims to cover the context of the target domain with the given models.
Although the PoseNet trained with these datasets produces plausible results on real images, we observed that it does not detect a consistent number of instances across viewpoints of a single scene. This detection mismatch problem causes incorrectly generated weakly-paired images: some object instances are rendered in the wrong pose or not at all.
% false-positives to generate weakly-paired images. False-positives 
%Detection mismatches cause , resulting in the generation of incorrectly paired images.
Since images are taken from various viewpoints in a shared scene (see~\autoref{data_collection}) , they should contain the same number of detections. Thus, we eliminate images that do not match that number. This is called \textit{mismatch filter}.
We first run inference on every view $i$ in a scene and compute the number of detections for a given object, $n_i$ in a range of distances (in our case, from 0.3\rm{m} to 1.5\rm{m}).
% , covering common use-cases (in our case, from 0.3\rm{m} to 1.5\rm{m}).
We then compute which value of $n$ occurs most frequently, $\bar{n}$, and only use the views where $n_i = \bar{n}$ for further process.
For example, if inferred results have one instance in 90\% of images, two instances in 9\%, and zero instances in 1\%, we eliminate the latter 10\% of images.
Finally, we render synthetic images based on the estimated poses and obtain weakly-paired images.
% We qualitatively observe that 3D bounding boxes match very well the object location, as seen in \autoref{fig:sim2real_overview}'s (2)(3). However, since our real images are not labeled, we cannot run a quantitative evaluation.
% The projected 3D bounding box in image space are almost identical, as can be seen in \autoref{fig:sim2real_overview}'s (2)(3). Insert here accuracy at <5px error: 80%
% However, the 6D pose error is comparatively larger, 6D pose accuracy at 5cm 5 deg : 75%,  since the small pixel-level errors are amplified by the deprojection (or PnP) procedure.
% While some errors of 6D pose remain, the object pixel locations between synthetic and real images are nearly identical, as can be seen in \autoref{fig:sim2real_overview}'s (2)(3).

% accuracy at <5px error: 80%
% 6D pose accuracy at 5cm 5 deg : 75%
% https://arxiv.org/abs/1711.10006

\subsection{Instance-level Style Transfer}  \label{style_transfer}
% object appearance = intrinsic appearance + style
% intrinsic appearance = all features needed to distinguish pose and class of object. For example, shape and texture. In a renderer, that would be geometry + material (Physically Based Material) for a new, standard instance.
% style = the rest (Lighting, shadows, superficial variations in texture)
% intrinsic appearance
Style transfer must preserve the shape and texture details of each target object. This is important since these features are entangled with the object's 6D pose.
In this section, we show how to transfer the target object's style from synthetic to real images, using weakly-paired images.

\begin{figure}[t]
\centering
\includegraphics[scale=0.30]{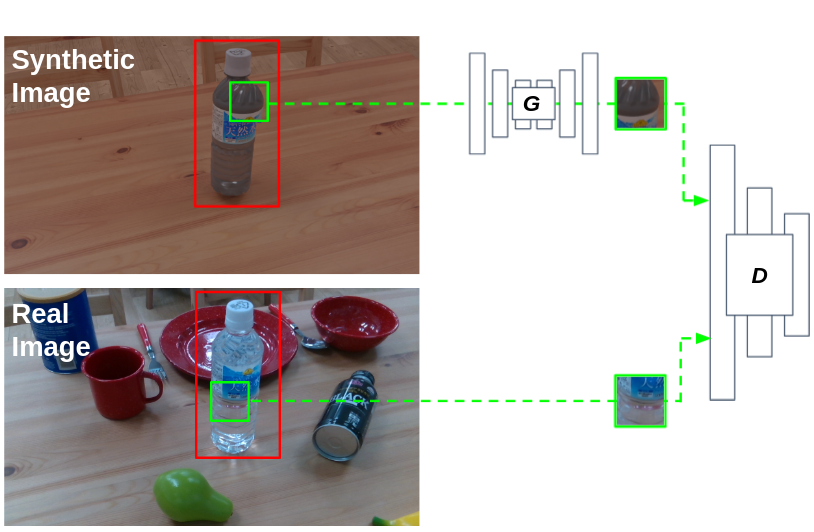}
\caption{Training of an instance-level style transfer network on weakly-paired images. $G$ represents the style-mapping function from synthetic to real. $D$ represents the discriminator for synthetic and real.}
\label{fig:style_transfer}
\end{figure}

Our TransNet is inspired from single image translations as described in CUT \cite{Park2020-lf}. This method partially satisfies the above requirements by controlling the patch size for generator and discriminator.
However, to solve the proper mapping from source to target, they assume that the images have similar contents in both domains (see Fig.~9 in \cite{Park2020-lf}). While these assumptions allow the method to work, it is difficult to satisfy these in general indoor scenes.
On the other hand, because weakly-paired images already contain similar contents around the object area between sim and real, we leverage these in order to eliminate the need for this assumption (see \autoref{fig:style_transfer}).
The main difference between the original and our proposed algorithm is how to process the input and output image. The details are described below. 

For TransNet training, we first crop both the synthetic and real images with the 2D bounding box of the target object area, using synthetic mask labels. In our case, the input image size is 640 x 480 and we extract a cropped area 1.2 times larger than the 2D bounding box. 
Then, we randomly scale the cropped images between 170 to 512 on the longest axis while keeping the aspect ratio. % major axes. 
% Lastly, we randomly crop each scaled image as a patch of 64 x 64. 
Lastly, we randomly sample fixed-size patches of 64 x 64 from the cropped-and-scaled image. % This final crop
This sampling is carried out independently for synthetic and real images. % The cropped images 
The patches are used as the network input (see \autoref{fig:style_transfer}). Using small patches of the target area allows us to preserve structure which is key of style transfer for PoseNet.
While the original image transformation \cite{Park2020-lf} includes random flipping, we have removed it. This is because asymmetrical information, such as labels and lettering, should be kept in our case. 
In addition, although the original method extracts input information from only a single image on each domain, we can use multiple images since all images are weakly-paired.
Instead of a single \textit{image} translation, our network performs a single \textit{object style} translation. In other words, we train one TransNet for each object class, using multiple images.

The network architecture and loss function are the same as the single image translation \cite{Peng2019-mk}.
The loss function is defined as: 
\begin{align*}
L_{G} =&\ L_{patchNCE}(G,H,X) + L_{patchNCE}(G,H,Y) +\\
    &\ L_{GAN}(G,D,X,Y) + L_{1}(G,Y)\\
L_{D} =&\ L_{GAN}(G,D,X,Y) + R_{1}(D,Y)  
\end{align*}
where $X$ and $Y$ represent the synthetic and real images, respectively; $G$ represents the style-mapping function from synthetic to real; $D$ is the style discriminator for synthetic and real; and $H$ is a two-layer multilayer perceptron (MLP).
The Generator and Discriminator are based on StyleGAN2 \cite{Karras2021-xz}. The Generator loss function is a combination of PatchNCE and non-saturating GAN Loss \cite{radford2015unsupervised} with L1 regularization. The Discriminator loss is non-saturating GAN Loss with R1 gradient penalty stabilization \cite{Karras2021-xz, Mescheder2018-it}.
For more detail, please see the original paper \cite{Park2020-lf}'s appendix B.

\begin{figure}[t]
\centering
\includegraphics[scale=0.21]{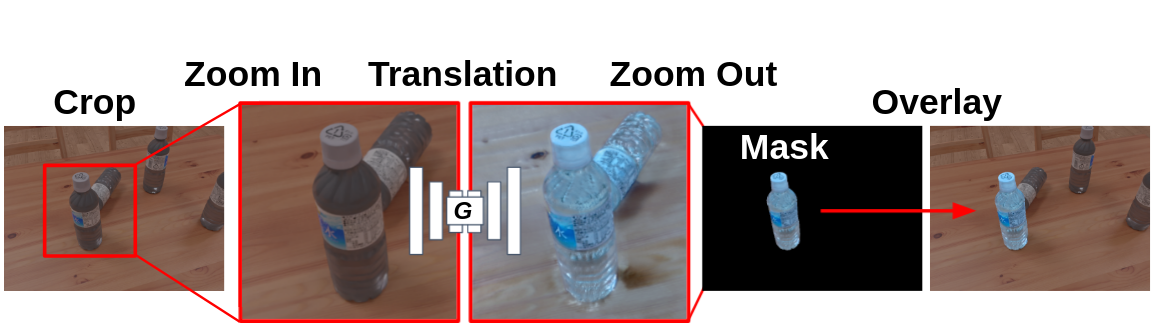}
\caption{The process of individual object style transfer}
% Flow of style transfer around a target object area}
\label{fig:inference}
\end{figure}

For style transfer with sim2real, we would like to preserve the object shape without losing details. Therefore, we carry out the following pre-process and post-process as shown in \autoref{fig:inference}.
Before translation, we crop the input synthetic image with a 2D square bounding box and enlarge the image, in our case we scaled it to 512 x 512.
% After this process, we transfer the enlarged synthetic image and crop it with the masked area. This operation has the benefit of keeping the object silhouette. Lastly, we re-scale the image to its original size. This operation tends to result in less blurry images after translation (see appendix \ref{resolution}).
After this process, we transfer the enlarged synthetic image and re-scale the image to its original size. This operation tends to result in less blurry images after translation (see appendix \ref{resolution}). Lastly, we crop it with the masked area and overlay the masked image to the input image. This operation has the benefit of keeping the object silhouette.
The following \autoref{alg:inference_flow} presents the entire inference flow for a single image:
\begin{algorithm}
\begin{algorithmic}[1]
    \REQUIRE image, network weights for target objects
    \SetKwComment{Comment}{$\triangleright$\ }{}
    \SetKwFunction{Infer}{Infer}
    \SetKwFunction{Mask}{Mask}
    \SetKwFunction{Crop}{Crop}
    \SetKwFunction{Resize}{Resize}
    \SetKwFunction{Overlay}{Overlay}
    %\STATE{$x \gets \Infer_0(image)$} \Comment*[r]{for background}
    \STATE{$x \gets image$}
    \FOR{$i \gets \text{class in image}$}
        \FOR{$j \gets \text{instance in class}$}
            \STATE{$x_j \gets \Crop(image)$}
            \STATE{$x_j \gets \Resize(x_j)$}
            \STATE{$x_j \gets \Infer_i(x_j)$}
            \STATE{$x_j \gets \Resize(x_j)$}
            \STATE{$x_j \gets \Mask(x_j)$}
            \STATE{$x \gets \Overlay(x, x_j)$}
        \ENDFOR
    \ENDFOR
    \STATE{$\textbf{return} \ x$}
\end{algorithmic}
\caption{Inference flow for a single image}
\label{alg:inference_flow}
\end{algorithm}

% Firstly, we use one arbitrary TransNet to transfer the entire image, without the above processing, to be used as the background.
Firstly, we load an input image to be used as the background.
Then, we conduct the style transfer for each instance using the corresponding TransNet.  
Lastly, all of the transferred instance images are overlaid on the background image.
%Interestingly, we see that any of the trained TransNets can be utilized for the style transfer of the background. This is because the cropped training images contain part of the background, as shown in \autoref{fig:style_transfer}.

\subsection{Training a 6D Pose Estimation Network} \label{training6D}
After training a TransNet for each object, you can transfer synthetic images that contains any number and any combination of target objects in one scene. For example, you can transfer an image which contains 5 water bottles, even if you trained the TransNet with only 1 water bottle in a scene, as shown in \autoref{fig:motivation}.
Once the synthetic data has been adapted by our TransNet, the PoseNet is trained.

% \begin{figure*}[!ht]
% \centering
% \includegraphics[scale=0.45]{images/dataset_eval.png}
% \caption{Examples of training datasets for PoseNet evaluation: (a) synthetic dataset. (b) synthetic dataset with our adaptation.}
% \label{fig:eval_data}
% \end{figure*}

\begin{figure*}[!ht]
\centering
\includegraphics[scale=0.29]{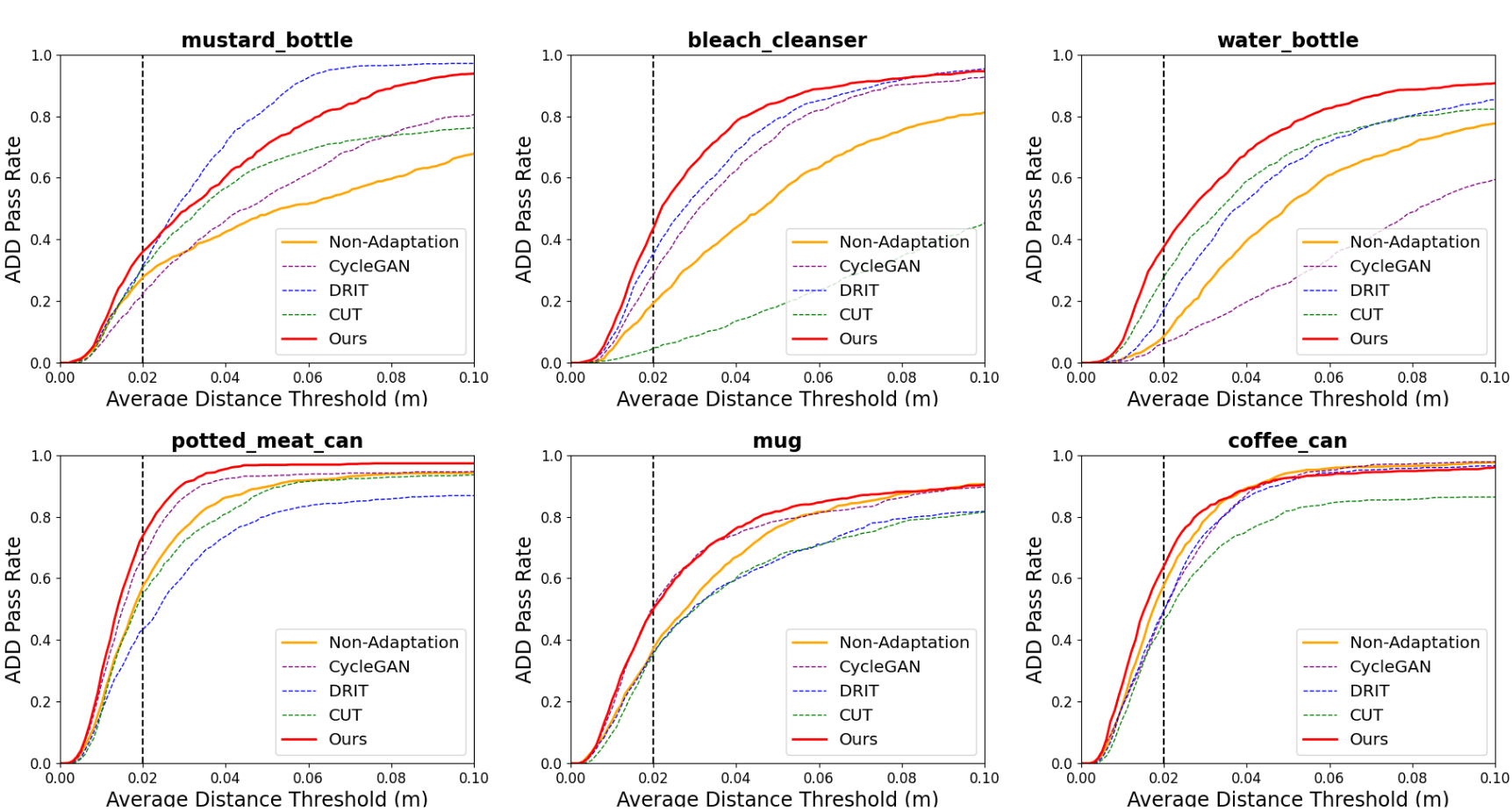}
\caption{% Average distance threshold curves of our adaptation method and non-adaptation for 6 household objects. %CPGs. 
Average distance threshold curves of non-adaptation, our adaptation method, and other methods including CycleGAN \cite{Zhu2017-mq}, DRIT \cite{lee2020drit++}, CUT \cite{Park2020-lf}.}
% The score of area under the curve (AUC) is shown in the legend.}
\label{fig:add_curve}
\end{figure*}

\begin{table*}[ht]
% \begin{center}
\centering
\caption{ADD pass rate of 2cm threshold and Score of area under the curve (AUC) for 6 household objects}
\label{tab:add_score} 
% \begin{tabular}{ |p{2cm}||p{0.75cm}p{0.75cm}|p{0.75cm}p{0.75cm}|p{0.75cm}p{0.75cm}|p{0.75cm}p{0.75cm}|p{0.75cm}p{0.75cm}|p{0.75cm}p{0.75cm}|}
\begin{tabular}{ |p{2cm}||p{0.78cm}p{0.78cm}|p{0.78cm}p{0.78cm}|p{0.78cm}p{0.78cm}|p{0.78cm}p{0.78cm}|p{0.78cm}p{0.78cm}|p{0.78cm}p{0.78cm}|}
 \hline
 \multirow{2}{*}{\textbf{Data / Object}} & \multicolumn{2}{c|} {mustard\_bottle} & \multicolumn{2}{c|} {potted\_meat\_can} & \multicolumn{2}{c|} {bleach\_cleanser} & \multicolumn{2}{c|} {mug} & \multicolumn{2}{c|} {water\_bottle} & \multicolumn{2}{c|}{coffee\_can}\\
 \cline{2-3} \cline{4-5} \cline{6-7} \cline{8-9} \cline{10-11} \cline{12-13} 
 & ADD & AUC & ADD & AUC & ADD & AUC & ADD & AUC & ADD & AUC & ADD & AUC\\
 \hline \hline
 Non-adaptation   & 0.280 & 42.74 & 0.579 & 74.50 & 0.196 & 48.32 & 0.377 & 63.39 & 0.089 & 43.62 & 0.584 & 76.99\\
 \hline
 CycleGAN \cite{Zhu2017-mq} & 0.227 & 48.67 & 0.674 & 78.38 & 0.290 & 61.51 & \textbf{0.512} & 66.97 & 0.063 & 27.41 & 0.503 & 75.19\\
 \hline
 DRIT \cite{lee2020drit++} & 0.321 & \textbf{67.77} & 0.437 & 65.63 & 0.357 & 65.12 & 0.364 & 57.16 & 0.176 & 52.64 & 0.499 & 74.56\\
 \hline
 CUT \cite{Park2020-lf} & 0.312 & 53.14 & 0.554 & 72.59 & 0.049 & 19.57 & 0.358 & 56.41 & 0.283 & 55.63 & 0.469 & 66.49\\
 \hline
 Ours   & \textbf{0.363} & 61.61 & \textbf{0.742} & \textbf{81.75} & \textbf{0.442} & \textbf{69.33} & 0.505 & \textbf{68.42} & \textbf{0.381} & \textbf{63.70} & \textbf{0.645} & \textbf{77.65}\\
 \hline 
\end{tabular}
% \end{center}
\end{table*}

\section{Experiments} \label{experiments}
In this section, we quantitatively evaluate the effect of sim2real adaptation using our TransNets on the performance of PoseNets.
% as shown in \autoref{fig:add_curve} and \autoref{tab:add_score}. 
Quantitative and qualitative comparison against several popular style transfer networks is also conducted.

\subsection{Experimental environment}

\begin{figure}[t]
\centering
\includegraphics[scale=0.18]{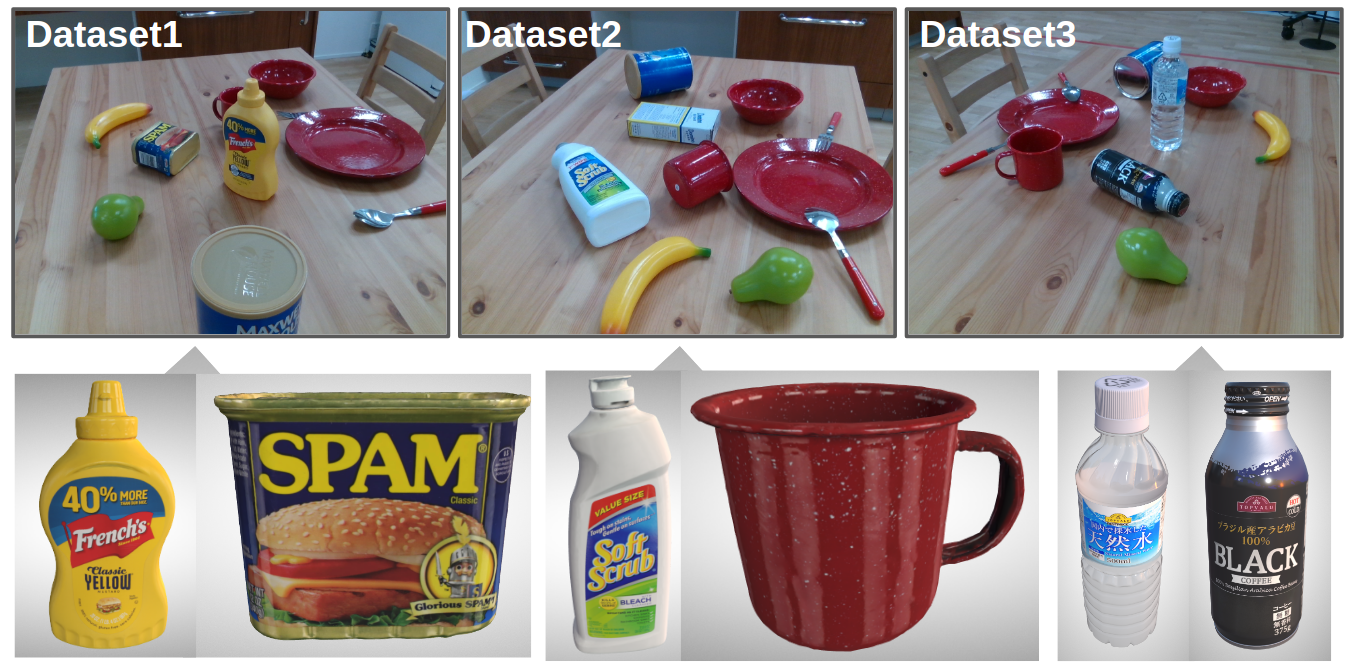}
\caption{\textbf{Target environment and objects:} The mustard\_bottle and potted\_meat\_can are target objects in Dataset1. The bleach\_cleanser and mug are targets in Dataset2. The water\_bottle and coffee\_can are targets in Dataset3. All other objects are distractors.}
\label{fig:env_setting}
\end{figure}

Since we are targeting indoor environments in the real world, we evaluated our approach on a dining table in a home, as shown in \autoref{fig:env_setting}. 
Our robot is installed on an X-Y gantry, above the table, and is able to access various viewing angles of a given object on the tabletop.
For our target objects, we selected two sets of household objects. %CPGs.
The first set is the popular \textit{YCB Object Set} \cite{calli2015benchmarking}. We picked 4 objects from this item set, representing a variety of colors and shapes: 
006\_mustard\_bottle, 010\_potted\_meat\_can, 021\_bleach\_cleanser, and 025\_mug.
Since %CPGs
household objects often contain transparent and shiny objects, we aim to evaluate our approach using objects with those features. Unfortunately, the YCB objects only contain base color information, and lack shading data. Thus, we have added two Japanese household %CPG 
object models, containing physically realistic and challenging material properties: a transparent water\_bottle and a shiny coffee\_can, as shown in \autoref{fig:env_setting}.

\begin{figure*}[!ht]
\centering
\includegraphics[scale=0.47]{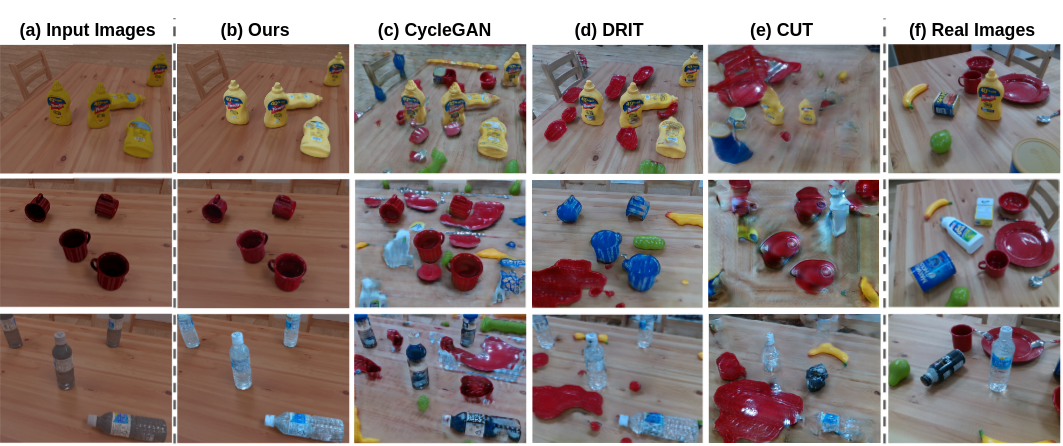}
\caption{\textbf{Style Transfer Results}: We compare ours to other transfer methods including CycleGAN \cite{Zhu2017-mq}, DRIT \cite{lee2020drit++}, CUT \cite{Park2020-lf}.}
\label{fig:da_comparison}
\end{figure*}

% \begin{figure*}[!ht]
% \centering
% % \includegraphics[scale=0.47]{images/add_comparison.png}
% \includegraphics[scale=0.58]{images/da_add.png}
% \caption{Average distance threshold curves of our adaptation method and CycleGAN \cite{Zhu2017-mq}, DRIT \cite{lee2020drit++}, CUT \cite{Park2020-lf}.}
% \label{fig:add_curve}
% \end{figure*}

\subsection{Datasets and Metrics}  \label{eval_data}

We created three real-world datasets for our experiments, as shown in \autoref{fig:env_setting}. Each dataset contains two target objects plus distractors (target objects: mustard\_bottle / potted\_meat\_can, bleach\_cleanser / mug, and water\_bottle / coffee\_can).
Target objects were randomly placed on the table. For each dataset, 1k RGB images were gathered at a range of distances from 0.3\rm{m} to 1.5\rm{m}, covering common use-cases. The images were captured by an \textit{Intel RealSense D415} mounted on the head of our robot, as shown in \autoref{fig:sim2real_overview} (0). 

We evaluated the 6D pose estimation performance using the Average Distance of Distinguishable Model Points (ADD) metric from \cite{xiang2017posecnn, Hinterstoisser2013-rf}. Ground truth labels, used only for evaluation, were obtained via manual labeling.

\subsection{Training}
We selected PVNet \cite{Peng2019-mk} as a PoseNet for our experiments. For the training, we used two types of synthetic datasets, as described in \autoref{weakly-paired}. These datasets contain a total of 5k images of each target object:
\begin{itemize}
  \item 2.5k Domain-Randomized images, as shown in \autoref{fig:sim2real_overview} (a). % \rm{\hspace{.18em}i\hspace{.18em}}.  %$\alpha$.
  \item 2.5k Domain-Specific images, as shown in \autoref{fig:sim2real_overview} (a).% \rm{\hspace{.08em}ii\hspace{.08em}}. % $\beta$.
\end{itemize}
Each image contains a maximum of 5 instances from the same class. We trained 6 PoseNets individually, one for each object, from scratch. % \autoref{fig:eval_data} 
\autoref{fig:da_comparison} (a) shows samples of the Domain-Specific images and \autoref{fig:da_comparison} (b) shows the results of our domain adaptation via TransNet. 
Since TransNet performs domain-specific adaptation, we only adapted Domain-Specific images; the Domain-Randomized dataset was used without domain adaptation.
% We only performed TransNet on the domain-specific dataset. 
For the PoseNet training, we set the epoch and batch sizes to 50 and 2 respectively.

For TransNet, we updated the data processing parts of single image translation \cite{Park2020-lf} to the one described in \autoref{style_transfer}. We gathered 1k real images for each dataset as described in the \autoref{eval_data}, and then sifted them using our mismatch filter. % false-positive filter.
The remainder (the average: 86.5\% in our case) was used for TransNet training, for which we set the epoch and batch sizes to 16 and 16 respectively.
% 903/1000, 921/1000, 783/1000, 905/1000, 830/1000, 853/1000 images are left for 6 objects.

All networks were implemented using \textit{PyTorch}.
% and trained using \textit{AWS p3 instance}. 
%\textit{Kubeflow} is used to build the entire pipeline. 
For rendering synthetic images, we used \textit{Blender Cycles}.

\subsection{Evaluation Results and Comparison}

% To exhibit how PoseNet is adapted to a target domain via our TransNet, we performed a comparison, using the ADD metric, with our method and 3 other methods: CycleGAN  \cite{Zhu2017-mq}, DRIT \cite{lee2020drit++}, and CUT \cite{Park2020-lf}.
To exhibit how PoseNet is adapted to a target domain via our TransNet, we performed a comparison between non-adaptation and our adaptation, using the ADD metric. Moreover, we compared these and 3 other methods: CycleGAN  \cite{Zhu2017-mq}, DRIT \cite{lee2020drit++}, and CUT \cite{Park2020-lf}.
To ensure a fair comparison we used the same data in all cases and trained the same PoseNet on the adapted images. Although the compared methods can accept unpaired images, we trained them using the same weakly paired images used for our TransNet. 
% The compared methods can not transfer domain-randomized images, those are excluded.
% The same datasets used for training our TransNet are also used for training the other methods. To perform a fair comparison, the images for training are loaded in pairs and only domain-specific images are adapted via these methods for PoseNet training. % after the training.
% each network is given only domain-specific images which are then transferred to realistic images.
\autoref{fig:add_curve} shows the resulting average distance threshold curves: the orange solid curve shows the accuracy of PoseNet trained on the synthetic data without adaptation, the red solid curve shows with our adaptation, the purple dashed curve shows with CycleGAN \cite{Zhu2017-mq}, the blue dashed curve shows with DRIT \cite{lee2020drit++}, the green dashed curve shows with CUT \cite{Park2020-lf}. 
% The orange curve represents the accuracy of the PoseNet trained on the synthetic dataset without adaptation. The red curve represents the PoseNet trained on the dataset with our adaptation. 
The vertical dotted line indicates a 2\rm{cm} threshold. This level of accuracy is necessary in order to accomplish proficient object grasping using a robot, as described in DOPE \cite{tremblay2018deep}.
% To perform a fair comparison, the training images are loaded in pairs for all networks.

% The results in \autoref{fig:add_curve} and \autoref{tab:add_score}
%indicate that our TransNets improved the performance of all PoseNets in the target domain. 
The results in \autoref{fig:add_curve} and \autoref{tab:add_score} 
% and \autoref{tab:add_score}
indicate that only our TransNets improved the performance for all target objects compared to the non-adaptation. In contrast, the all other works show uneven results.
% Explain that related works have more variance?
For example, while DRIT \cite{lee2020drit++} successfully improved the AUC score for mustard\_bottle, the scores for potted\_meat\_can, mug and coffee\_can are lower than non-adaptation. 
% CUT \cite{Park2020-lf} failed to improve the performance for mug. CycleGAN \cite{Zhu2017-mq} failed to improve the performance for water\_bottle. 
They also show that our approach is applicable not only to opaque objects, but also transparent.
%In terms of the mean score, the proposed method is the best.
% of all PoseNets in the target domain. 

\autoref{fig:da_comparison} presents a qualitative comparison between our method and the above 3 methods.
% For a qualitative evaluation of our TransNet, \autoref{fig:da_comparison} presents results obtained with our method and 3 other methods: CycleGAN  \cite{Zhu2017-mq}, DRIT \cite{lee2020drit++}, and CUT \cite{Park2020-lf}.
% a qualitative  evaluation  of  our  TransNet,  as  shown  in Fig. 7, compares results obtained with our method and those of  3  other  methods:
% To perform a fair comparison, all data is generated in pairs for all training scenarios (one each for our method and a competing method).
% To perform a fair comparison, the training images are loaded in pairs for all networks.
The competing methods contain collapsing shapes (the second row of \autoref{fig:da_comparison} (e)), style mismatches (the second row of \autoref{fig:da_comparison} (d)), and visible artifacts around objects (the third row of \autoref{fig:da_comparison} (c)). 
The quality of these adapted images is therefore unreliable depending on scene complexity. We hypothesize that these effects, for example, the style mismatch of DRIT on the mug (the second row of \autoref{fig:da_comparison} (d)), are responsible for its poor 6D pose estimation performance on that object.
In contrast, our method is able to generate images stably without these effects. 
% We hypothesize that the style mismatches of DRIT on the mug, CycleGAN on the water\_bottle, and the shape collapse of CUT, etc are responsible for its poor performance on that object.  

% even in complex scenes.
% These networks attempt to generate distractor objects that were present in the training data but not at inference time. In contrast, our method does not overfit to the training data in this manner.

\subsection{Discussion}

% \begin{figure}[!ht]
% \centering
% \includegraphics[scale=0.56]{images/discussion4.png}
% \caption{\textbf{Sim2Real Gaps}: (a) The appearance gap is small. (b) Synthetic image is sub-optimal due to transparent material settings. (c) The model is different from the real one which only has the red at the top of its front label. % The label of the modeldiffers from that of the real one. 
% (d) The gap is large due to shiny material settings. Our approach is capable of bridging these gaps. % bridging all of the above gaps
% }
% \label{fig:discussion}
% \end{figure}

\begin{figure}[!ht]
\centering
\includegraphics[scale=0.56]{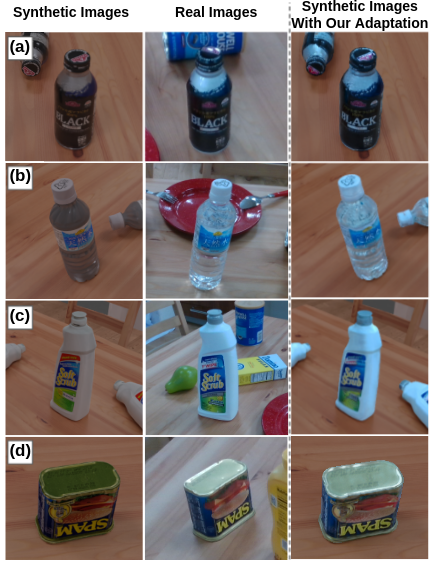}
\caption{\textbf{Sim2Real Gaps}: (a) The appearance gap is small. (b) Synthetic image is sub-optimal due to transparent material settings. (c) The model is different from the real one which only has the red at the top of its front label. % The label of the modeldiffers from that of the real one. 
(d) The gap is large due to shiny material settings. Our approach is capable of bridging these gaps. % bridging all of the above gaps
}
\label{fig:discussion}
\end{figure}

% TODO(taku): update DRIT results
% \begin{figure*}[!ht]
% \centering
% %\includegraphics[scale=0.46]{images/da_comparison.png}
% \includegraphics[scale=0.28]{images/da_comparison5.png}
% \caption{\textbf{Style Transfer Results}: We compare ours to other transfer methods including CycleGAN \cite{Zhu2017-mq}, DRIT \cite{Lee2018-db}, CUT \cite{Park2020-lf}.}
% \label{fig:da_comparison}
% \end{figure*}
% In one case, 
% Our approach produces consistent results even if the gap between synthetic and real domains is big. Indeed, 
% We have noticed that our approach benefits most in cases where there is a significant gap between sim and real. 
We have noticed that our approach benefits where there is a significant gap between sim and real.
% We have noticed that the effectiveness of our approach is proportional to the appearance gap in sim2real.
On one hand, the texture/material quality of the 3D-modeled coffee\_can shown in \autoref{fig:discussion} (a) is already excellent. As the domain gap is small, the improvement is trivial (an increased score of AUC: 0.66).
On the other hand, big gains are had for the two objects that have large domain gaps in modeling: water\_bottle (with an increased score of AUC: 20.08) and bleach\_cleanser (with an increased score of AUC: 21.01). As shown in \autoref{fig:discussion} (b), the synthetic image of water\_bottle is very different from the real one, mainly due to the complexity of modeling and rendering transparent materials. In the case of bleach\_cleanser, our real object does not have a red label that's present on the 3D model due to a change in packaging as shown in \autoref{fig:discussion} (c). It is quite common for the appearance of household objects % CPGs 
(labeling details, textures, text, etc.) to change. For example, seasonal accents might be introduced, promotional labels added, or expiration date placement and details changed. Since our approach is capable of bridging these gaps successfully, it will enable wider usage of sim2real. 
% On the other hand, the 3D model of the coffee\_can is already high quality as shown in \autoref{fig:discussion} (c). In that case, the improvement is small Since the domain gap is already small.

\section{Conclusions}
% Note(taku): 結論の説は過去形
We showed a sim2real domain adaptation technique for 6D pose estimation, without any manual annotations.
Our approach was able to automatically generate weakly-paired synthetic and real datasets that are traditionally difficult to obtain.
% The proposed method was able to translate synthetic images to those which were close to target domain images, as shown in \autoref{fig:discussion}.
The proposed method was able to translate the styles of texture/material-rich objects from synthetic to realistic in cluttered indoor scenes, as shown in \autoref{fig:env_setting} and \autoref{fig:discussion}.
Moreover, evaluation results proved the proposed method amplifies the performance of a 6D pose estimation network for all target objects. As such, when 3D models and camera poses are available, one can automatically adapt the network to a target environment in the real world at low cost. 

In this paper, we have focused on the style transfer of the foreground only. One of the next steps will be to perform background style transfer.
% As a next step, we will tackle the style transfer of the background.
% Furthermore, while we assumed 3D models are given, in some cases it might be difficult to assume it. 
% Furthermore, we shouldn't assume that 3D models will always be available.
% Also, one of the key inputs to our methods is any given object's 3D model. However, a 3D model cannot always be assumed. 
Also, while we assumed 3D models are given, that is not always the case.
Therefore, in future works, we would like to employ real2sim techniques that automatically produce 3D models from real images, such as photogrammetry and NeRF \cite{Mildenhall2020-hv}.
% In this time, while we assumed 3D models are given, in some cases it might be difficult to assume it. For future works, we would like to employ real2sim techniques that automatically reconstruct models from real images, such as photogrammetry, NeRF \cite{Mildenhall2020-hv}, etc. 

\appendix
% \section*{APPENDIX}

\subsection{Scaling Operation for TransNet Inference}  \label{resolution}

\begin{figure}[htbp]
\centering
\includegraphics[scale=0.43]{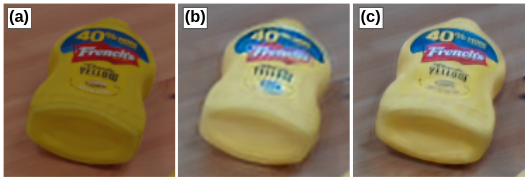}
\caption{(a) Input image. (b) Transferred image without scale-up. (c) Transferred image with scale-up.}
\label{fig:resolution}
\end{figure}

The scaling operation as described in \autoref{style_transfer} preserves texture details, as shown in \autoref{fig:resolution}. Indeed, the text of the label in \autoref{fig:resolution} (b) without the scaling operation is blurry.

% \subsection{Style Transfer of Shiny Material}  \label{specular}

% Figure.~\ref{fig:shiny} shows one style transfer result of a shiny material which is the bottle of potted\_meat\_can. The appearance of the synthetic image tend to be different since YCB models don't have rich-information to represent gloss.
% \begin{figure}[h]
% \centering
% \includegraphics[scale=0.33]{images/shiny_material.png}
% \caption{Style transfer result of a shiny material (the bottom of potted\_meat\_can)}
% \label{fig:shiny}
% \end{figure}

% \subsection{Domain-Randomized Dataset}  \label{dr-dataset}

% To generate domain-randomized datasets as shown in \autoref{fig:sim2real_overview} $\alpha$ and \autoref{fig:dr_dataset},
% we randomized object poses, distractor types (primitives and SuperShapes) and textures (solid, striped, and image), point light positions and intensities, and background images from the MS COCO dataset \cite{Lin2014-gv}.

% \begin{figure}[htbp]
% \centering
% \includegraphics[scale=0.33]{images/dr_dataset.png}
% \caption{Examples of domain-randomized images}
% \label{fig:dr_dataset}
% \end{figure}

\subsection{Style Transfer for Multi-Class}  \label{multi-class inference}

\begin{figure*}[t]
\centering
\includegraphics[scale=0.58]{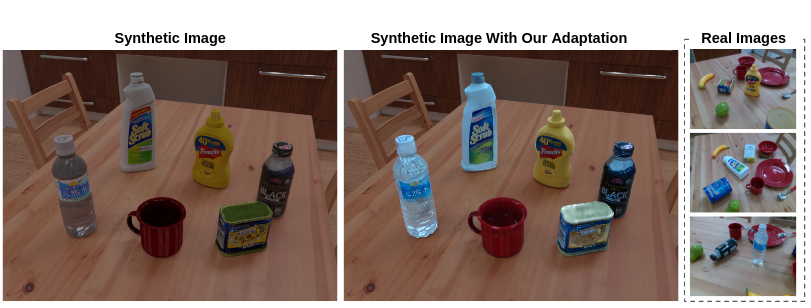}
\caption{Style transfer results for multi-class objects in a shared scene}
\label{fig:multiclass}
\end{figure*}

% Fig.~\ref{fig:resolution} shows inference results of TransNets for multi-class in a image. 
Once you train individual TransNets for each target object, you can transfer synthetic images which contain multiple classes, as shown in \autoref{fig:multiclass}. Since training is done independently for each object, style transfer can be completed without training data containing all the target objects in the same scene. At inference time, individual instances of each class are cropped out, as shown in \autoref{fig:inference}, and inference is performed with the TransNet corresponding to that class.

\bibliographystyle{IEEEtran}
\bibliography{IEEEabrv,myrefs}

\end{document}